\documentclass[letterpaper]{article}
\usepackage{aaai}
\usepackage{times}
\usepackage{helvet}
\usepackage{courier}
\usepackage{amssymb}
\usepackage{amsmath}
\usepackage{enumerate}
\usepackage{epsf}
\usepackage{algorithm}
\usepackage{wrapfig}
\usepackage{sidecap}
\usepackage{color}
\usepackage{multirow}
\usepackage{rotating}
\usepackage{lscape}
\usepackage{graphicx}
\usepackage{subfigure}
\usepackage[english]{babel}
\usepackage{multicol}
\usepackage{amsmath,amsfonts,amsthm,amssymb}
\usepackage{algpseudocode}
\usepackage{footnote}
\usepackage{graphicx,subfigure}
\usepackage{times}
\newcommand{\data}{\mathcal{D}}
\newcommand{\ep}{\mathbb{E}}
\newcommand{\R}{\mathcal{R}}

\newcommand{\xv}{\mathbf{x}}

\newcommand{\Vv}{\mathbf{V}}

\newcommand{\wv}{\mathbf{w}}

\newcommand{\risk}{\mathcal{R}}

\newcommand{\lambdav}{\boldsymbol \lambda }
\newcommand{\thetav}{\boldsymbol{\theta}}

\newtheorem{lem}{Lemma}

\begin{document}
%

\title{Dropout Training for Support Vector Machines}
\author{Ning Chen~~~~ Jun Zhu~~~~ Jianfei Chen~~~~ Bo Zhang\\
State Key Lab of Intelligent Tech. \& Systems; Tsinghua National TNList Lab;\\
Department of Computer Science and Technology, Tsinghua University, Beijing 100084, China\\
\{ningchen@mail, dcszj@mail, chenjf10@mails, dcszb@mail\}.tsinghua.edu.cn}
\maketitle

\begin{abstract}
\begin{quote}
Dropout and other feature noising schemes have shown promising results in controlling over-fitting by artificially corrupting the training data. Though extensive theoretical and empirical studies have been performed for generalized linear models, little work has been done for support vector machines (SVMs), one of the most successful approaches for supervised learning. This paper presents dropout training for linear SVMs. To deal with the intractable expectation of the non-smooth hinge loss under corrupting distributions, we develop an iteratively re-weighted least square (IRLS) algorithm by exploring data augmentation techniques. Our algorithm iteratively minimizes the expectation of a re-weighted least square problem, where the re-weights have closed-form solutions. The similar ideas are applied to develop a new IRLS algorithm for the expected logistic loss under corrupting distributions. Our algorithms offer insights on the connection and difference between the hinge loss and logistic loss in dropout training. Empirical results on several real datasets demonstrate the effectiveness of dropout training on significantly boosting the classification accuracy of linear SVMs.
\end{quote}
\end{abstract}

\section{Introduction}

Artificial feature noising augments the finite training data with an infinite number of corrupted versions, by corrupting the given training examples with a fixed noise distribution. Among the many noising schemes, dropout training~\cite{Hinton:dropout} is an effective way to control over-fitting by randomly omitting subsets of features at each iteration of a training procedure. By formulating the feature noising methods as minimizing the expectation of some loss functions under the corrupting distributions, recent work has provided theoretical understandings of such schemes from the perspective of adaptive regularization~\cite{PercyLiang13}; and has shown promising empirical results in various applications, including document classification~\cite{CorruptICML2013,PercyLiang13}, named entity recognition~\cite{Wang:emnlp13}, and image classification~\cite{Wang:icml13}.

Regarding the loss functions, though much work has been done on the quadratic loss, logistic loss, or the log-loss induced from a generalized linear model (GLM)~\cite{CorruptICML2013,PercyLiang13,Wang:emnlp13}, little work has been done on the margin-based hinge loss underlying the very successful support vector machines (SVMs)~\cite{Vapnik:95}. One technical challenge is that the non-smoothness of the hinge loss makes it hard to compute or even approximate its expectation under a given corrupting distribution. Existing methods are not directly applicable, therefore calling for new solutions. This paper attempts to address this challenge and fill up the gap by extending dropout training as well as other feature noising schemes to support vector machines.

Previous efforts on learning SVMs with feature noising have been devoted to either explicit corruption or an adversarial worst-case analysis. For example, virtual support vector machines~\cite{Burges1997} explicitly augment the training data, which are usually support vectors from previous learning iterations for computational efficiency, with additional examples that are corrupted through some invariant transformation models. A standard SVM is then learned on the corrupted data. Though simple and effective, such an approach lacks elegance and the computational cost of processing the additional corrupted examples could be prohibitive for many applications. The other work~\cite{Globerson:icml06,Dekel:icml08,Teo2008} adopts an adversarial worst-case analysis to improve the robustness of SVMs against feature deletion in testing data. Though rigorous in theory, a worst-case scenario is unlikely to be encountered in practice. Moreover, the worst-case analysis usually results in solving a complex and computationally demanding problem.


In this paper, we show that it is efficient to train linear SVM predictors on an infinite amount of corrupted copies of the training data by marginalizing out the corruption distributions, an average-case analysis. We concentrate on dropout training, but the results are directly applicable to other noising models, such as Gaussian, Poisson and Laplace~\cite{CorruptICML2013}. For all these noising schemes, the resulting expected hinge loss can be upper-bounded by a variational objective by introducing auxiliary variables, which follow a generalized inverse Gaussian distribution. We then develop an iteratively re-weighted least square (IRLS) algorithm to minimize the variational bounds. At each iteration, our algorithm minimizes the expectation of a re-weighted quadratic loss under the given corrupting distribution, where the re-weights are computed in a simple closed form.
We further apply the similar ideas to develop a new IRLS algorithm for the dropout training of logistic regression, which extends the well-known IRLS algorithm for standard logistic regression~\cite{Hastie:2009}. Our IRLS algorithms shed light on the connection and difference between the hinge loss and logistic loss in the context of dropout training, complementing to the previous analysis~\cite{Rosasco:2004,Globerson:07} in the supervised learning settings. Finally, empirical results on classification and a challenging ``nightmare at test time" scenario~\cite{Globerson:icml06} demonstrate the effectiveness of our approaches, in comparison with various strong competitors.


\section{Preliminaries}

We setup the problem in question and review the learning with marginalized corrupted features.

\subsection{Regularized loss minimization}

Consider the binary classification, where each training example is a pair $(\xv, y)$ with $\xv \in \mathbb{R}^D$ being an input feature vector and $y \in \{+1, -1\}$ being a binary label. Given a set of training data $\data = \{(\xv_n, y_n)\}_{n=1}^N$, supervised learning aims to find a function $f \in \mathcal{F}$ that maps each input to a label. To find the optimal candidate, it commonly solves a regularized loss minimization problem
\begin{eqnarray}
\min_{f \in \mathcal{F}}~ \Omega(f) + 2c \cdot \risk(\data; f),
\end{eqnarray}
where $\risk(\data; f)$ is the risk of applying $f$ to the training data; $\Omega(f)$ is a regularization term to control over-fitting; and $c$ is a non-negative regularization parameter.

For linear models, the function $f$ is simply parameterized as $f(\xv; \wv, b) = \wv^\top \xv + b$, where $\wv$ is the weight vector and $b$ is an offset. We will denote $\thetav :=\{\wv, b\}$ for clarity. Then, the regularization can be any Euclidean norms\footnote{It is a common practice to not regularize the offset.}, e.g., the $\ell_2$-norm, $\Omega(\wv) = \Vert \wv \Vert_2^2$, or the $\ell_1$-norm, $\Omega(\wv) = \Vert \wv \Vert_1$. For the loss functions, the most relevant measure is the training error, $ \sum_{n=1}^N \delta( f(\xv_n; \thetav) \neq y_n)$, which however is not easy to optimize. A convex surrogate loss is used instead, which normally upper bounds the training error. Two popular examples are the hinge loss and logistic loss\footnote{The natural logarithm is not an upper bound of the training error. We can simply change the base without affecting learning.}:\\[-.7cm]

{\small \setlength\arraycolsep{1pt} \begin{eqnarray}
\risk_h( \data; \thetav ) &=&   \sum_{n=1}^N \max\left(0, \ell - y_n f(\xv_n; \thetav)\right), \nonumber \\
\risk_l( \data; \thetav ) &=&   \sum_{n=1}^N \left(-\log p(y_n | \xv_n, \thetav) \right), \nonumber
\end{eqnarray}}\\[-.7cm]

\noindent where $\ell (>0)$ is the cost of making a wrong prediction, and $p(y_n | \xv_n, \thetav) := 1 / (1 + \exp(-y_n f(\xv_n; \thetav) ) )$ is the logistic likelihood. Other losses include the quadratic loss, $\sum_{n=1}^N (f(\xv_n; \thetav) - y_n)^2$, and the exponential loss, $\sum_{n=1}^N \exp( - y_n f(\xv_n; \thetav) )$, whose feature noising analyses are relatively simpler~\cite{CorruptICML2013}.

\subsection{Learning with marginalized corruption}

Let $\tilde{\xv}$ be the corrupted version of the input features $\xv$. Consider the commonly used independent corrupting model:
$$p(\tilde{\xv} | \xv )=\prod_{d=1}^D p(\tilde{x}_{d} | x_d; \eta_d),$$
where each individual distribution is a member of the exponential family, with the natural parameter $\eta_d$. Another common assumption is that the corrupting distribution is unbiased, that is, $\ep_p[\tilde{\xv}  | \xv ] = \xv$, where we use $\ep_p[\cdot] := \ep_{p(\tilde{\xv}|\xv)}[\cdot]$ to denote the expectation taken over the corrupting distribution $p(\tilde{\xv} | \xv)$. Such examples include the unbiased blankout (or dropout) noise, Gaussian noise, Laplace noise, and Poisson noise~\cite{Vincent:2008,CorruptICML2013}.

For the {\it explicit corruption} in~\cite{Burges1997}, each example $(\xv_n, y_n)$ is corrupted $M$ times from the corrupting model $p(\tilde{\xv}_n|\xv_n)$, resulting in the corrupted examples $(\tilde{\xv}_{nm}, y_n)$, $m \in [M]$. This procedure generates a new corrupted data set $\tilde{\data}$ with a larger size of $NM$. The generated dataset can be trained by minimizing the average loss function over $M$ corrupted data points:\\[-.7cm]

{\small \begin{eqnarray}\label{eq:ExplicitCorruption}
\mathcal{L}(\tilde{\data}; \thetav)=\sum_{n=1}^N \frac{1}{M} \sum_{m=1}^{M}\mathcal{R}(\tilde{\xv}_{nm}, y_n; \thetav),
\end{eqnarray}}\\[-.7cm]

\noindent where $\mathcal{R}(\xv, y; \thetav)$ is the loss function of the model incurred on the training example $(\xv, y)$. As $\mathcal{L}(\tilde{\data}; \thetav)$ scales linearly with the number of corrupted observations, this approach may suffer from high computational costs.

Dropout training adopts the strategy of {\it implicit corruption}, which learns the model with marginalized corrupted features by minimizing the expectation of a loss function under the corrupting distribution\\[-.7cm]

{\small \begin{eqnarray}\label{eq:ExpectedLoss}
\mathcal{L}(\data; \thetav) = \sum_{n=1}^{N}\ep_{p}[ \R(\tilde{\xv}_n, y_n; \thetav)].
\end{eqnarray}}\\[-.7cm]

\noindent The objective can be seen as a limit case of~(\ref{eq:ExplicitCorruption}) when $M \to \infty$, by the law of large numbers. Such an expectation scheme has been widely adopted in previous work~\cite{PercyLiang13,CorruptICML2013,Wang:emnlp13,Wang:icml13}.

The choice of the loss function $\mathcal{R}$ in (\ref{eq:ExpectedLoss}) can make a significant difference, in terms of computation cost and prediction accuracy. Previous work on feature noising has covered the quadratic loss, exponential loss, logistic loss, and the loss induced from generalized linear models (GLM). For the quadratic loss and exponential loss, the expectation in Eq.~(\ref{eq:ExpectedLoss}) can be computed analytically, thereby leading to simple gradient descent algorithms~\cite{CorruptICML2013}. However, it does not have a closed form to compute the expectation for the logistic loss or the GLM loss. Previous analysis has resorted to approximation methods, such as using the second-order Taylor expansion~\cite{PercyLiang13} or an upper bound by applying Jensen's inequality~\cite{CorruptICML2013}, both of which lead to effective algorithms in practice. In contrast, little work has been done on the hinge loss, for which the expectation under corrupting distributions cannot be analytically computed either, therefore calling for new algorithms.


\section{Learning SVMs with Corrupting Noise}

We now present a simple iteratively re-weighted least square (IRLS) algorithm to learn SVMs with the expected hinge loss under corrupting distributions. Our method consists of a variational upper bound of the expected loss and a simple algorithm that iteratively minimizes an expectation of a re-weighted quadratic loss. We also apply the similar ideas to develop a simple IRLS algorithm for minimizing the expected logistic loss, thereby allowing for a systematical comparison of the hinge loss with the logistic and quadratic losses in the context of feature noising.

\subsection{A variational bound with data augmentation}

Let $\zeta_n := \ell - y_n(\wv^\top\tilde{\xv}_n)$.\footnote{We treat the offset $b$ implicitly by augmenting $\xv_n$ and $\tilde{\xv}_n$ with one dimension of deterministic $1$. More details will be given in the algorithm.} Then, the expected hinge loss can be written as
\begin{eqnarray}\label{eqn:EpHingeLoss}
\R_h(\data; \thetav) = \sum_{n=1}^N \ep_{p}[\max{(0, \zeta_n)}],
\end{eqnarray}
Since we do not have a closed-form of the expectation of the max function, minimizing the expected loss~(\ref{eqn:EpHingeLoss}) is intractable. Here, we derive a variational upper bound based on a data augmentation formulation of the expected hinge loss. Let $\phi(y_n|\tilde{\xv}_n, \thetav) = \exp\{-2c \max(0, \zeta_n)\}$  be the pseudo-likelihood of the response variable for sample $n$. Then we have
\begin{eqnarray}
\risk_h(\data; \thetav) = - \frac{1}{2c} \sum_n \ep_p [ \log \phi(y_n | \tilde{\xv}_n, \thetav) ].
\end{eqnarray}
Using the ideas of data augmentation~\cite{Polson:BA11,Zhu:jmlr14}, the pseudo-likelihood can be expressed as
\setlength\arraycolsep{-3pt} \begin{eqnarray}\label{eqn:lemmaScaleMixture}
&& \phi(y_n | \tilde{\xv}_n, \thetav) = \int_0^\infty \!\!\! \frac{1}{\sqrt{2\pi\lambda_n}}\exp\left\{-\frac{(\lambda_n+c\zeta_n)^2}{2\lambda_n}\right\}d\lambda_n,
\end{eqnarray}
where $\lambda_n,~n \in [N]$, is the augmented variable. Using (\ref{eqn:lemmaScaleMixture}) and Jensen's inequality, we can derive a variational upper bound $\mathcal{L}$ of the expected hinge loss as
\setlength\arraycolsep{1pt} \begin{eqnarray}\label{eqn:Upperbound}
\mathcal{L}(\thetav, q(\lambdav)) &=& \sum_{n=1}^N \Big\{-H(\lambda_n) + \frac{1}{2}\ep_q \lbrack \log\lambda_n\rbrack\\
&+& \ep_q \Big\lbrack\frac{1}{2\lambda_n}\ep_{p} (\lambda_n + c\zeta_n)^2\Big\rbrack \Big\} + \textrm{constant}, \nonumber
\end{eqnarray}
where $H(\lambda_n)$ is the entropy of the variational distribution $q(\lambda_n)$; $q(\lambdav) := \prod_n q(\lambda_n)$ is joint distribution; and we have defined $\ep_q[\cdot] := \ep_{q(\lambdav)}[\cdot]$ to denote the expectation taken over a variational distribution $q$. Now, our variational optimization problem is
\begin{eqnarray}\label{eq:variationalBound}
\min_{\thetav, q(\lambdav) \in \mathcal{P} } \Vert \wv \Vert_2^2 + \mathcal{L}(\thetav, q(\lambdav) ),
\end{eqnarray}
where $\mathcal{P}$ is the simplex space of normalized distributions. We should note that when there is no feature noise (i.e., $\tilde{\xv} = \xv$), the bound is tight and we are learning the standard SVM classifier. Please see Appendix A for the derivation. We will empirically compare with SVM in experiments.

\subsection{Iteratively Re-weighted Least Square Algorithm}

In the upper bound, we note that when the variational distribution $q(\lambdav)$ is given, the term $\ep_{p} [(\lambda_n + c\zeta_n)^2]$ is an expectation of a quadratic loss, which can be analytically computed. 
We leverage such a nice property and develop a coordinate descent algorithm to solve problem~(\ref{eq:variationalBound}). Our algorithm iteratively solves the following two steps, analogous to the common two-step procedure of a variational EM algorithm.

{\bf For $q(\lambdav)$ (i.e., E-step)}: infer the variational distribution $q(\lambdav)$. Specifically, optimize $\mathcal{L}$ over $q(\lambdav)$, we get:
\setlength\arraycolsep{1pt} \begin{eqnarray}
    q(\lambda_n) &\propto& \frac{1}{\sqrt{\lambda_n}}\exp\left\{-\frac{1}{2}\left( \lambda_n + \frac{c^2\ep_p [\zeta_n^2]}{\lambda_n} \right)\right\}\nonumber\\
    &\sim&\mathcal{GIG}\left( \lambda_n; \frac{1}{2}, 1, c^2\ep_{p}[\zeta_n^2] \right),
    \end{eqnarray}
    where the second-order expectation is
    \begin{eqnarray}
    \ep_p[\zeta_n^2] =&&  \wv^\top(\ep_p[\tilde{\xv}_n]\ep_p[\tilde{\xv}_n ]^\top  + V_p[\tilde{\xv}_n]) \wv \nonumber \\
    && - 2 \ell  y_n \wv^\top\ep_p[\tilde{\xv}_n] + \ell^2;
    \end{eqnarray}
    and $V_p[\tilde{\xv}_n]$ is a $D\times D$ diagonal matrix with the $d$th diagonal element being the variance of $\tilde{x}_{nd}$, under the corrupting distribution $p(\tilde{\xv}_n|\xv_n)$. We have denoted $\mathcal{GIG}(x;p,a,b) \propto x^{p-1}\exp(-\frac{1}{2}(\frac{b}{x}+ax))$ as a generalized inverse Gaussian distribution. 
    Thus, $\lambda_n^{-1}$ follows an inverse Gaussian distribution
    \begin{eqnarray}\label{eqn:inverseGaussian}
    q(\lambda_n^{-1}|\tilde{\xv}_n, \thetav) \sim \mathcal{IG}\left( \lambda_n^{-1}; \frac{1}{c\sqrt{\ep[\zeta_n^2]}}, 1 \right)
    \end{eqnarray}

    {\bf For $\thetav:= \wv$ (i.e., M-step)}: removing irrelevant terms, this step involves minimizing the following objective:
    \setlength\arraycolsep{-2pt} \begin{eqnarray}\label{eqn:Upperbound-w-b}
    && \mathcal{L}_{[\thetav]} 
                           = \Vert \wv \Vert_2^2  + \sum_{n=1}^N  \ep_{p}\left[c \zeta_n  +  \frac{c^2}{2} \gamma_n \zeta_n^2 \right],
    \end{eqnarray}
    where $\gamma_n := \ep_q[\lambda_n^{-1}]$. We observe that this substep is equivalent to minimizing the expectation of a re-weighted quadratic loss, as summarized in Lemma 1, whose proof is deferred to Appendix B, for brevity.
\begin{lem}
Given $q(\lambdav)$, the M-step minimizes the re-weighted quadratic loss (with the $\ell_2$-norm regularizer):
\begin{eqnarray}
\Vert \wv \Vert_2^2 + \frac{c^2}{2} \sum_n \gamma_n \ep_{p} \left[ (\wv^\top \tilde{\xv}_n - y_n^h)^2 \right],
\end{eqnarray}
where $y_n^h = (\ell + \frac{1}{c \gamma_n}) y_n$ is the re-weighted label, and the re-weights are computed in closed-form:
\begin{eqnarray}\label{eq:ExpLambda}
\gamma_n := \ep_{q}[\lambda_n^{-1}] 
= \frac{1}{c\sqrt{\ep_p[\zeta_n^2]}}.
\end{eqnarray}
\end{lem}
For low-dimensional data, we can solve for the closed form solutions by doing matrix inversion. Specifically, optimizing $\mathcal{L}_{[\thetav]}$ over $\wv$, we get\footnote{To consider offset, we simply augment $\xv$ and $\tilde{\xv}$ with an additional unit of $1$. The variance $V_p[\tilde{\xv}_n]$ is augmented accordingly. The identity matrix $I$ is augmented by adding one zero row and one zero column.}:
\setlength\arraycolsep{1pt} \begin{eqnarray}
    \wv &=& \left(\frac{2}{c^2} I +  \sum_{n=1}^N \gamma_n( \ep_p[ \tilde{\xv}_n \tilde{\xv}_n^\top]) \right)^{-1} \left(\sum_{n=1}^N \gamma_n y_n^h \ep_p[\tilde{\xv}_n] \right),  \nonumber
    \end{eqnarray}
where $\ep_p[ \tilde{\xv}_n \tilde{\xv}_n^\top ] =  \ep_p[\tilde{\xv}_n] \ep_p[\tilde{\xv}_n]^\top + V_p[\tilde{\xv}_n]$. However, if the data are in a high-dimensional space, e.g., text documents, the above matrix inversion will be computationally expensive. In such cases, we can use numerical methods, e.g., the quasi-Newton method.

To summarize, our algorithm iteratively minimizes the expectation of a simple re-weighted quadratic loss under the given corrupting distribution, where the re-weights $\gamma_n$ are computed in an analytic form. Therefore, it is an extension of the classical {\it iteratively re-weighted least square} (IRLS) algorithm~\cite{Hastie:2009} for dropout training. 
We also observe that if we fix $\gamma_n$ at $\frac{1}{c}$ and set $\ell=0$, we are minimizing the quadratic loss under the corrupting distribution, as studied in~\cite{CorruptICML2013}.
We will empirically show that our iterative algorithm for the expected hinge-loss will consistently improve over the standard quadratic loss by adaptively updating $\gamma_n$.
Finally, as we assume that the corrupting distribution is unbiased, i.e., $\ep_{p}[\tilde{\xv} | \xv ] = \xv$, we only need to compute the variance of the corrupting distribution, which is easy for all the existing exponential family distributions. An overview of the variance of the commonly used corrupting distributions can be found in \cite{CorruptICML2013}.

\subsection{An IRLS algorithm for the logistic Loss}

We now extend the above ideas to develop a new IRLS algorithm for the logistic-loss, which also minimizes the expectation of a re-weighted quadratic loss under the corrupting distribution and computes the re-weights analytically.

Let $\omega_n := \wv^\top \tilde{\xv}_n$.
Then the expected logistic loss under a corrupting distribution is
\begin{eqnarray}\label{eqn:EpLogistLoss}
\risk_l (\data; \wv) =  -\sum_{n=1}^N \ep_{p}\left[\log\left(\frac{e^{y_n \omega_n}}{1+e^{y_n\omega_n}}\right)\right].
\end{eqnarray}
Again since the expectation cannot be computed in closed-form, we derive a variational bound as a surrogate.
Specifically, let $\psi(y_n |\tilde{\xv}_n, \wv) = p^c(y_n | \tilde{\xv}_n, \wv)=\frac{e^{c y_n\omega_n}}{(1+e^{y_n\omega_n})^c}$ be the pseudo-likelihood of the response variable for sample $n$.
We have $\risk_l(\data; \wv) = - \frac{1}{c} \sum_n \ep_p[ \log \psi(y_n | \tilde{\xv}_n, \wv) ]$. Using the recent work of data augmentation~\cite{Polson:arXiv12,Chen:IJCAI2013}, the pseudo-likelihood can be expressed as
\setlength\arraycolsep{-4pt} \begin{eqnarray}\label{eqn:PolyaGammaEquality}
&& \psi(y_n | \tilde{\xv}_n, \wv)  = \frac{1}{2^c} e^{\kappa_n \omega_n}\int_0^\infty e^{-\frac{\lambda_n(y_n\omega_n)^2}{2}}p(\lambda_n)d\lambda_n,
\end{eqnarray}
where $\kappa_n := \frac{c}{2}y_n$ and $\lambda_n$ is the augmented Polya-gamma variable, $p(\lambda_n)\sim \mathcal{PG}(\lambda_n; c, 0)$. Using (\ref{eqn:PolyaGammaEquality}), we can derive the upper bound of the expected logistic loss:
\setlength\arraycolsep{1pt} \begin{eqnarray}
\mathcal{L}^\prime(\wv, q(\lambdav)) &=& \sum_{n=1}^N \Big\{\frac{1}{2}\ep_q[\lambda_n]\ep_p[\omega_n^2] - H(\lambda_n) \\
&& -\ep_q[\log p(\lambda_n)] -\frac{c}{2}y_n \ep_p[\omega_n ] \Big\} + \textrm{constant},\nonumber
\end{eqnarray}
and get the variational optimization problem
\begin{eqnarray}
\min_{\wv, q(\lambdav) \in \mathcal{P}} \Vert\wv\Vert^2_2 + \mathcal{L}^\prime(\wv, q(\lambdav)),
\end{eqnarray}
where $q(\lambdav)$ is the variational distribution


\begin{table}[t]\vspace{-.1cm}
\caption{Comparison of hinge loss and logistic loss under the IRLS algorithmic framework.}\label{table:lossComparison}
\begin{center}
 \scalebox{1}
 { \setlength{\tabcolsep}{1.8pt}
       \begin{tabular}{|c|c|c|c|c|}
       \hline
        \hline
           {}         & Parameter $\ell$ & Parameter $c$ & Update $\gamma_n$ & Update $y_n$ \\
        \hline
        Hinge       & $\ell$  & $c$ & Eq.~(\ref{eq:ExpLambda}) & $y_n^h$\\
        Logistic    & --  & $c$ & Eq.~(\ref{eq:ExpLambda2}) &  $y_n^l$ \\
        \hline
        \end{tabular}
}
\end{center}\vspace{-.3cm}
\end{table}

We solve the variational problem with a coordinate descent algorithm as follows:

{\bf For $q(\lambdav)$ (i.e., E-step)}: optimizing $\mathcal{L}^\prime$ over $q(\lambdav)$, we have:
\begin{eqnarray}
q(\lambda_n) &\propto& \exp\left(-\frac{1}{2}\lambda_n \ep_p[\omega_n^2]\right)p(\lambda_n | c,0)\nonumber\\
&\sim &\mathcal{PG}\left( \lambda_n; c, \sqrt{\ep_p[\omega_n^2]} \right)
\end{eqnarray}
a Polya-Gamma distribution~\cite{Polson:arXiv12}, where $\ep_p[\omega_n^2]=\wv^\top(\ep_p[\tilde{\xv}_n]\ep_p[\tilde{\xv}_n]^\top + V_p[\tilde{\xv}_n])\wv$. 

{\bf For $\wv$ (i.e., M-step)}: removing irrelevant terms, this step minimizes the objective
\setlength\arraycolsep{-3pt} \begin{eqnarray}
&& \mathcal{L}_{[\wv]}^\prime = \Vert\wv\Vert_2^2 + \sum_{n=1}^N \frac{1}{2}\ep_q[\lambda_n] \ep_p[\omega_n^2] - \frac{c}{2}y_n\ep_p[\omega_n].
\end{eqnarray}
We then have the optimal solution\footnote{The offset can be similarly incorporated as in the hinge loss.}:
\setlength\arraycolsep{1pt} \begin{eqnarray}
\wv &=& \left(I + \frac{1}{2}\sum_{n=1}^N \ep_q[\lambda_n] \ep_p[ \tilde{\xv}_n \tilde{\xv}_n^\top ] \right)^{-1} 
\left(\frac{c}{4}\sum_{n=1}^N y_n \ep_p[\tilde{\xv}_n]\right). \nonumber
\end{eqnarray}
This is actually equivalent to minimizing the expectation of a re-weighted quadratic loss, as in Lemma 2. The proof is similar to that of Lemma 1 and the expectation of a Polya-Gamma distribution follows~\cite{Polson:arXiv12}.
\begin{lem}
Given $q(\lambdav)$, the M-step minimizes the re-weighted quadratic loss (with the $\ell_2$-norm regularizer)
\begin{eqnarray}
\Vert \wv \Vert_2^2 + \frac{c}{2} \sum_n \gamma_n^l \ep_{p}[ (\wv^\top \tilde{\xv}_n - y_n^l)^2 ],
\end{eqnarray}
where $y_n^l = \frac{c}{2 \gamma_n} y_n$ is the re-weighted label, and $\gamma_n^l = \frac{\gamma_n}{c}$ with
\begin{eqnarray}\label{eq:ExpLambda2}
\gamma_n := \ep_{q}[\lambda_n] = \frac{c}{2\sqrt{\ep_p[\omega_n^2]}}\times\frac{e^{\sqrt{\ep_p[\omega_n^2]}}-1}{1+e^{\sqrt{\ep_p[\omega_n^2]}}}.
\end{eqnarray}
\end{lem}
It can be observed that if we fix $\gamma_n = \frac{c}{2}$, the IRLS algorithm reduces to minimizing the expected quadratic loss under the corrupting distribution. This is similar as in the case with SVMs, where if we set $\ell=0$ and fix $\gamma_n = \frac{1}{c}$, the IRLS algorithm for SVMs essentially minimizes the expected quadratic loss under the corrupting distribution.
Furthermore, by sharing a similar iterative structure, our IRLS algorithms shed light on the similarity and difference between the hinge loss and the logistic loss, as summarized in Table 1. Specifically, both losses can be minimized via iteratively minimizing the expectation of a re-weighted quadratic loss, while they differ in the update rules of the weights $\gamma_n$ and the labels $y_n$ at each iteration. 

\section{Experiments}
We now present empirical results on both classification and the challenging ``nightmare at test time" scenario~\cite{Globerson:icml06} to demonstrate the effectiveness of the dropout training algorithm for SVMs, denoted by Dropout-SVM, and the new IRLS algorithm for the dropout training of the logistic loss, denoted by Dropout-Logistic. We consider the unbiased dropout (or blankout) noise model\footnote{Other noising models (e.g., Poisson) were shown to perform worse than the dropout model~\cite{CorruptICML2013}. We have similar observations for Dropout-SVM and the new IRLS algorithm for logistic regression.}, that is,
$p(\tilde{\xv}=0)= q$ and $p(\tilde{\xv}=\frac{1}{1-q}\xv)= 1-q$, where $q \in [0,1)$ is a pre-specified corruption level. The variance of this model for each dimension $d$ is $V_p[\tilde{x}_d] = \frac{q}{1-q} x_d^2$.


\subsection{Binary classification}

We first evaluate Dropout-SVM and Dropout-Logistic on binary classification tasks. We use the public Amazon book review and kitchen review datasets~\cite{AmazonReviewData}, which consist of the text reviews about books and kitchen, respectively.
In both datasets, each document is represented as a 20,000 dimensional bag-of-words feature. The binary classification task is to distinguish whether a review content is positive or negative. Following the previous settings, we choose 2,000 documents for training and approximately 4,000 for testing.


We compare our methods with the methods presented in~\cite{CorruptICML2013} that minimize the quadratic loss with marginalized corrupted features (MCF), denoted by MCF-Quadratic, and that minimize the expected logistic loss, denoted by MCF-Logistic. MCF-Logistic was shown to be the state-of-the-art method for dropout training on these datasets, outperforming a wide range of competitors, including the dropout training of the exponential loss and the various loss functions with a Poisson noise model. As we have discussed, both Dropout-SVM and Dropout-Logistic iteratively minimize the expectation of a re-weighted quadratic loss, with the re-weights updated in closed-form. We include MCF-Quadratic as a baseline to demonstrate the effectiveness of our methods on adaptively tuning the re-weights to get improved results. We implement both Dropout-SVM and Dropout-Logistic using C++, and solve the re-weighted least square problems using L-BFGS methods~\cite{Liu:89}, which are very efficient by exploring the sparsity of bag-of-words features when computing gradients\footnote{We don't compare time with MCF methods, whose implementation are in Matlab (http://homepage.tudelft.nl/19j49/mcf/ Marginalized\_Corrupted\_Features.html), slower than ours.}.

\begin{figure}
\centering
\subfigure[books]{\includegraphics[height=1.6in, width=1.6in]{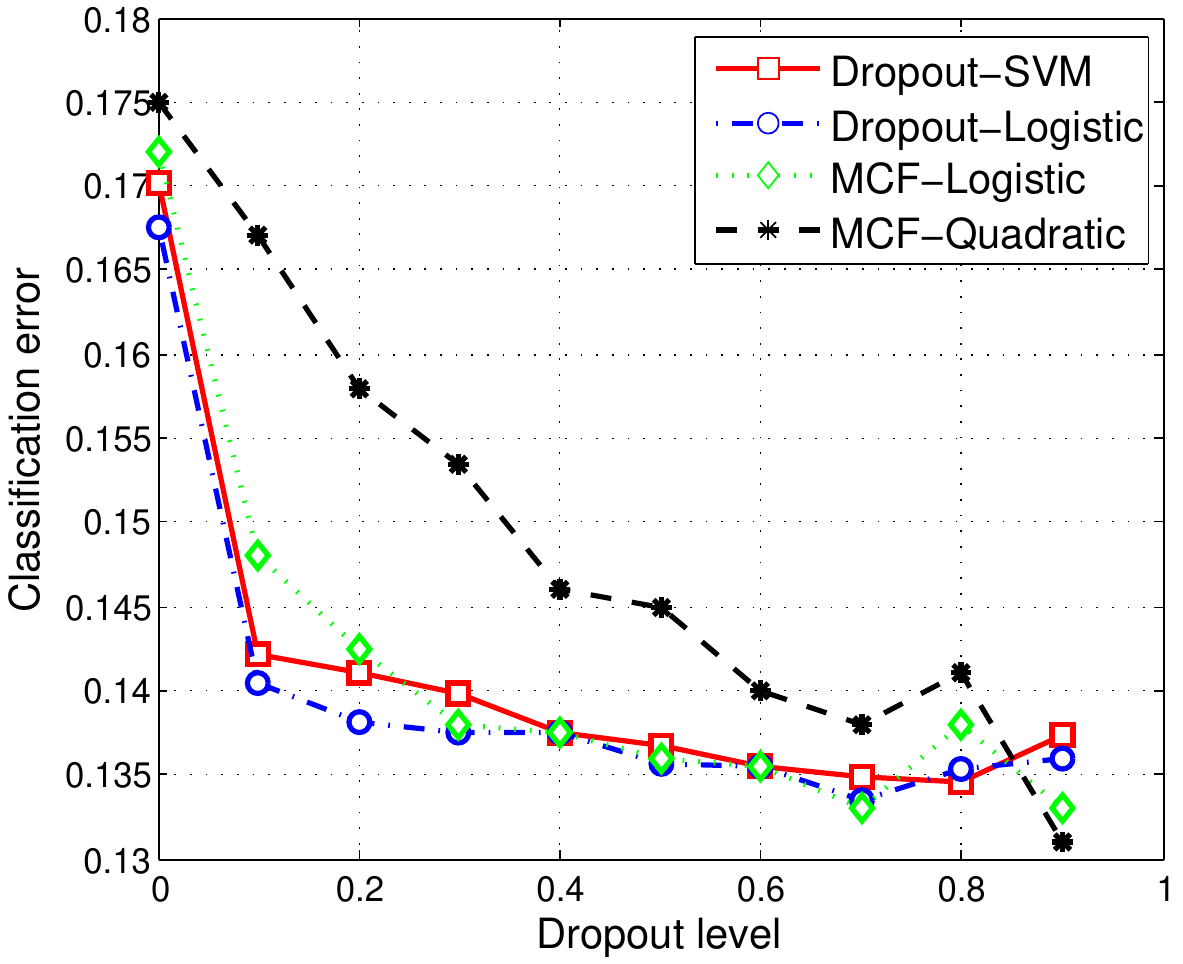}}\label{fig:amazon_books}\vspace{-.2cm}
\subfigure[kitchen]{\includegraphics[height=1.6in, width=1.6in]{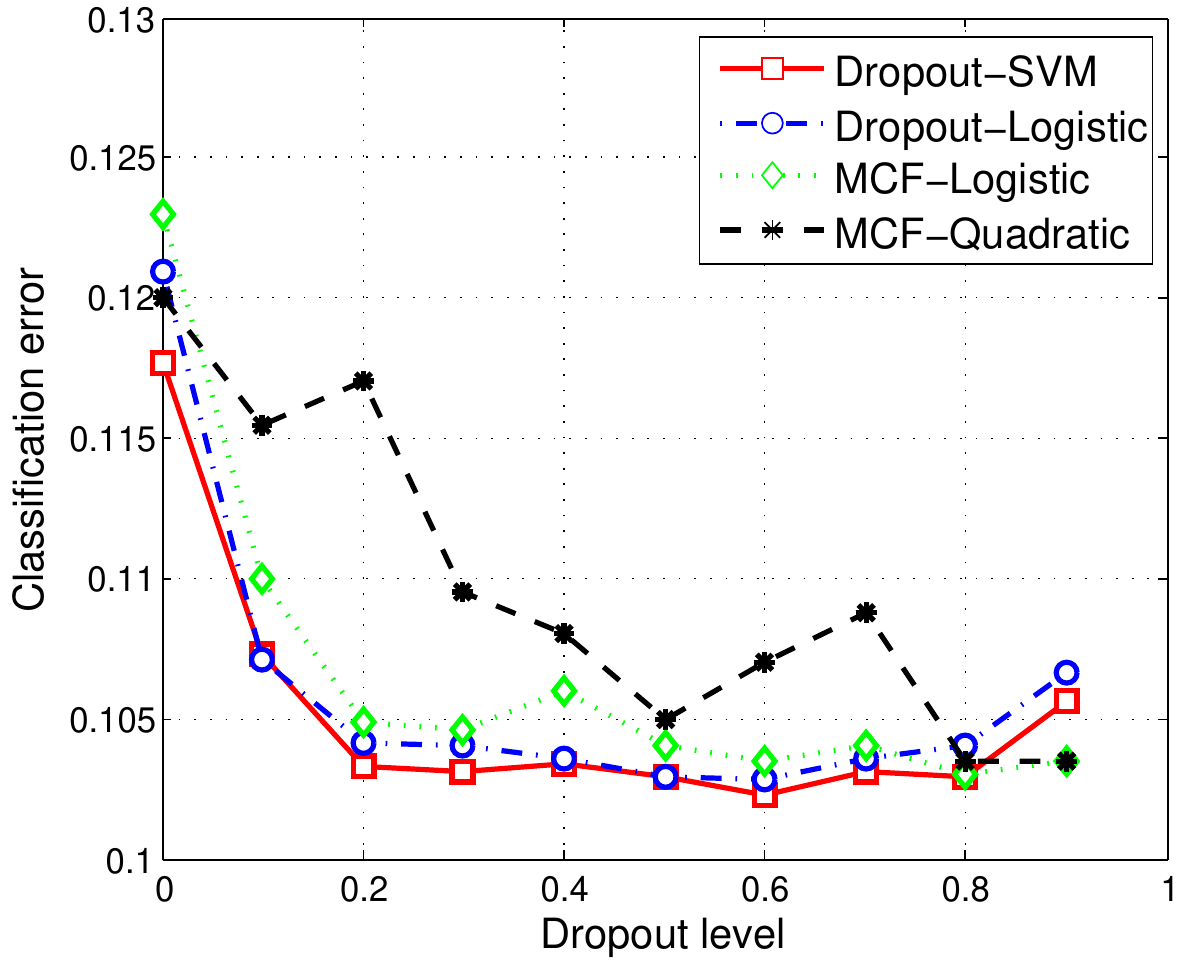}}\label{fig:amazon_kitchen}\vspace{-.2cm}
\caption{Classification errors on the Amazon datasets.}
\label{fig:amazon}\vspace{-.3cm}
\end{figure}

Figure~\ref{fig:amazon} shows classification errors, where the results of MCF-Logistic and MCF-Quadratic are cited from~\cite{CorruptICML2013}. We can see that on both datasets, Dropout-SVM and Dropout-Logistic generally outperform MCF-Quadratic except when the dropout level is larger than 0.9. In the meanwhile, the proposed two models give comparable results with (a bit better than on the kitchen dataset) the state-of-art MCF-Logistic which means that dropout training on SVMs is an effective strategy for binary classification. Finally, by noting that Dropout-SVM reduces to the standard SVM when the corruption level $q$ is zero, we can see that dropout training can significantly boost the classification performance for the simple linear SVMs.

\begin{figure}[h]\vspace{-.1cm}
\begin{center}
\includegraphics[width=.75\linewidth]{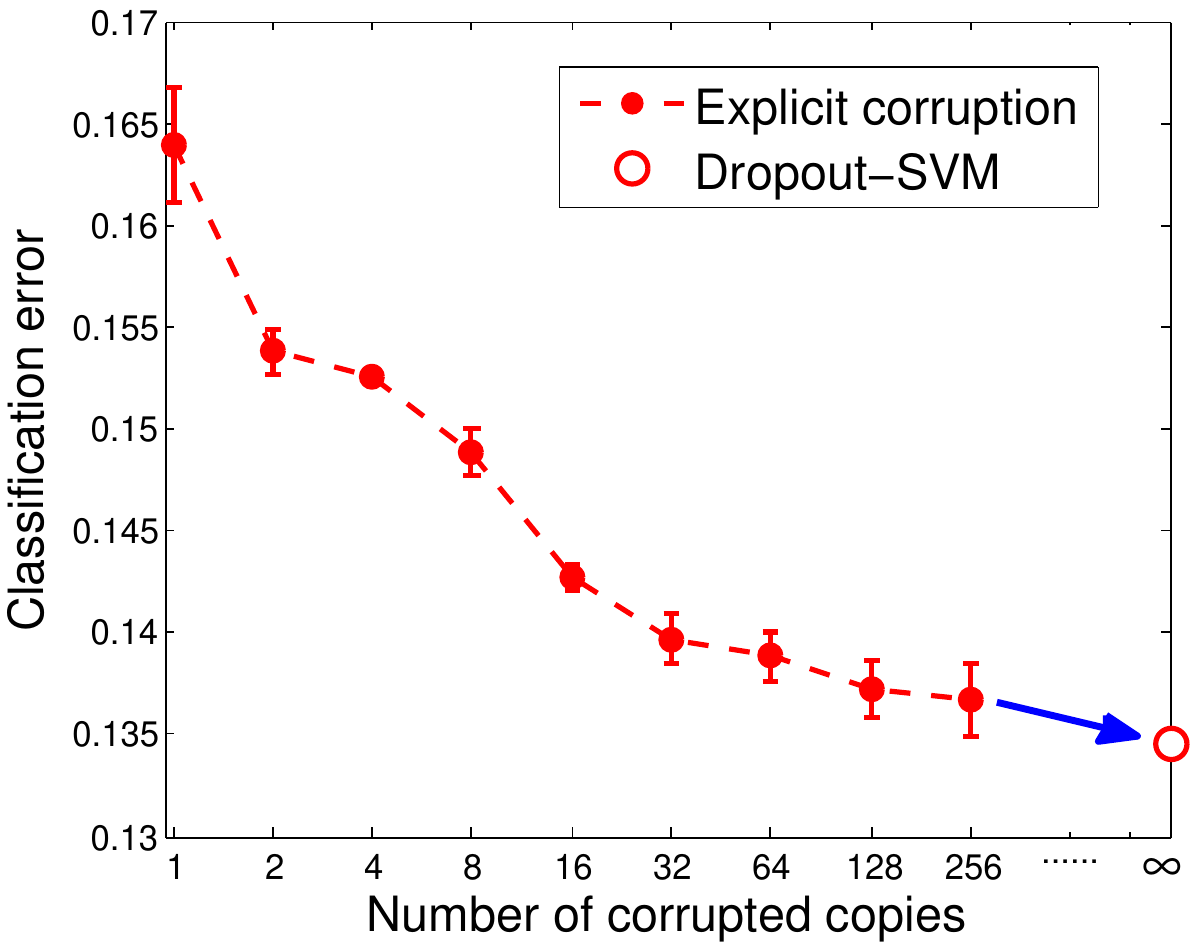}\vspace{-0.4cm}
\end{center}
\caption{Comparison between Dropout-SVM and the explicit corruption for SVM on the Amazon-books datasets.}
\label{fig:expSVM}\vspace{-0.4cm}
\end{figure}

\subsection{Dropout-SVM vs. Explicit corruption}

Figure~\ref{fig:expSVM} shows the classification errors on the Amazon-books dataset when a SVM classifier is trained using the explicit corruption strategy as in~Eq. (\ref{eq:ExplicitCorruption}). We change the number of corrupted copies (i.e., $M$) from $1$ to $256$. Following the previous setups~\cite{CorruptICML2013}, for each value of $M$ we choose the dropout model with $q$ selected by cross-validation. The hyper-parameter of the SVM classifier is also chosen via cross-validation on the training data. We can observe a clear trend that the error decreases when the training set contains more corrupted versions of the original training data, i.e., $M$ gets larger in Eq.~(\ref{eq:ExplicitCorruption}). It also shows that the best performance is obtained when $M$ approaches infinity, which is equivalent to our Dropout-SVM.


\subsection{Multi-class classification}

We also evaluate our methods on multiclass classification tasks. We choose the CIFAR-10 image categorization dataset\footnote{http://www.cs.toronto.edu/$\sim$kriz/cifar.html}. The CIFAR-10 dataset is the subset of the 80 million tiny images \cite{Torralba:PAMI08}. It consists of 10 classes of $32\times 32$ tiny images. We follow the experimental setup of the previous work~\cite{Cifar-10Data,CorruptICML2013} and represent each image as a 8,192 dimensional feature descriptor. We use the same 50,000 images for training and 10,000 for testing. There are various approaches to applying the binary Dropout-SVM and Dropout-Logistic to multiclass classification, including ``one-vs-all" and ``one-vs-one" strategies. Here we choose ``one-vs-all", which has shown effectiveness in many applications~\cite{InDefense:JMLR04}. The hyper-parameters are selected via cross-validation on the training set. 

Table~\ref{table:cifar10} presents the results, where the results of quadratic loss and logistic loss under the MCF learning setting\footnote{The exponential loss was shown to be worse; thus omitted.} are cited from \cite{CorruptICML2013}. We also report the results using Poisson noise. We can see that all the methods (except for the quadratic loss) can significantly boost the performance by adopting dropout training; meanwhile both Dropout-SVM and Dropout-Logistic are competitive, in fact achieving comparable performance as the state-of-the-art method (i.e., MCF-Logistic) under the dropout training setting. Finally, the Poisson corruption model is slightly worse than the dropout noise, consistent with the previous observations~\cite{CorruptICML2013}.
\begin{table}[t]\vspace{-.2cm}
\caption{Classification errors on CIFAR-10 data set.}\label{table:cifar10}
\begin{center}
 \scalebox{1.1}
 { \setlength{\tabcolsep}{1.8pt}
       \begin{tabular}{|c|c|c|c|}
       \hline
        \hline
        Model           &  No Corrupt  & Poisson & Dropout\\
        \hline
        Dropout-SVM        & 0.322 & 0.309 & 0.294 \\
        Dropout-Logistic     & 0.312 & 0.302 & 0.293\\
        MCF-Logistic        & 0.325 & 0.300 & 0.294\\
        MCF-Quadratic     & 0.326 & 0.291 & 0.323\\
        \hline
        \end{tabular}
}
\end{center}\vspace{-.5cm}
\end{table}

\subsection{Nightmare at test time}

Finally, we evaluate our methods under the ``nightmare at test time"~\cite{Globerson:icml06} supervised learning scenario, where some input features that were present when building the classifiers may ``die" or be deleted at testing time. In such a scenario, it is crucial to design algorithms that do not assign too much weight to any single feature during testing, no matter how informative it may seem at training. Previous work has conducted the worst-case analysis as well as the learning with marginalized corrupted features. We take this scenario to test the robustness of our dropout training algorithms for both SVM and logistic regression.

We follow the setup of \cite{CorruptICML2013}. Specifically, we choose the 
the MNIST dataset, which consists of 60,000 training and 10,000 testing handwritten digital images from 10 categories (i.e., $0, \cdots, 9$). The images are represented by $28\times 28$ pixels which results in the feature dimension of 784. We train the models on the full training set, and evaluate the performance on different versions of test set in which a certain level of the features are randomly dropped out, i.e., set to zero. We compare the performance of our dropout learning algorithms with the state-of-art MCF-predictors that use the logistic loss and quadratic loss. These two models also show the state-of-art performance on the same task to the best of our knowledge. We also compare with FDROP~\cite{Globerson:icml06}, which is a state-of-the-art algorithm for the ``nightmare at test time" setting that minimizes the hinge loss under an adversarial worst-case analysis. During training, we choose the best models over different dropout levels via cross-validation. For both Dropout-SVM and Dropout-Logistic, we adopt the ``one-vs-all" strategy as above for the multiclass classification task.

\begin{figure}
\centering
\includegraphics[width=.7\linewidth]{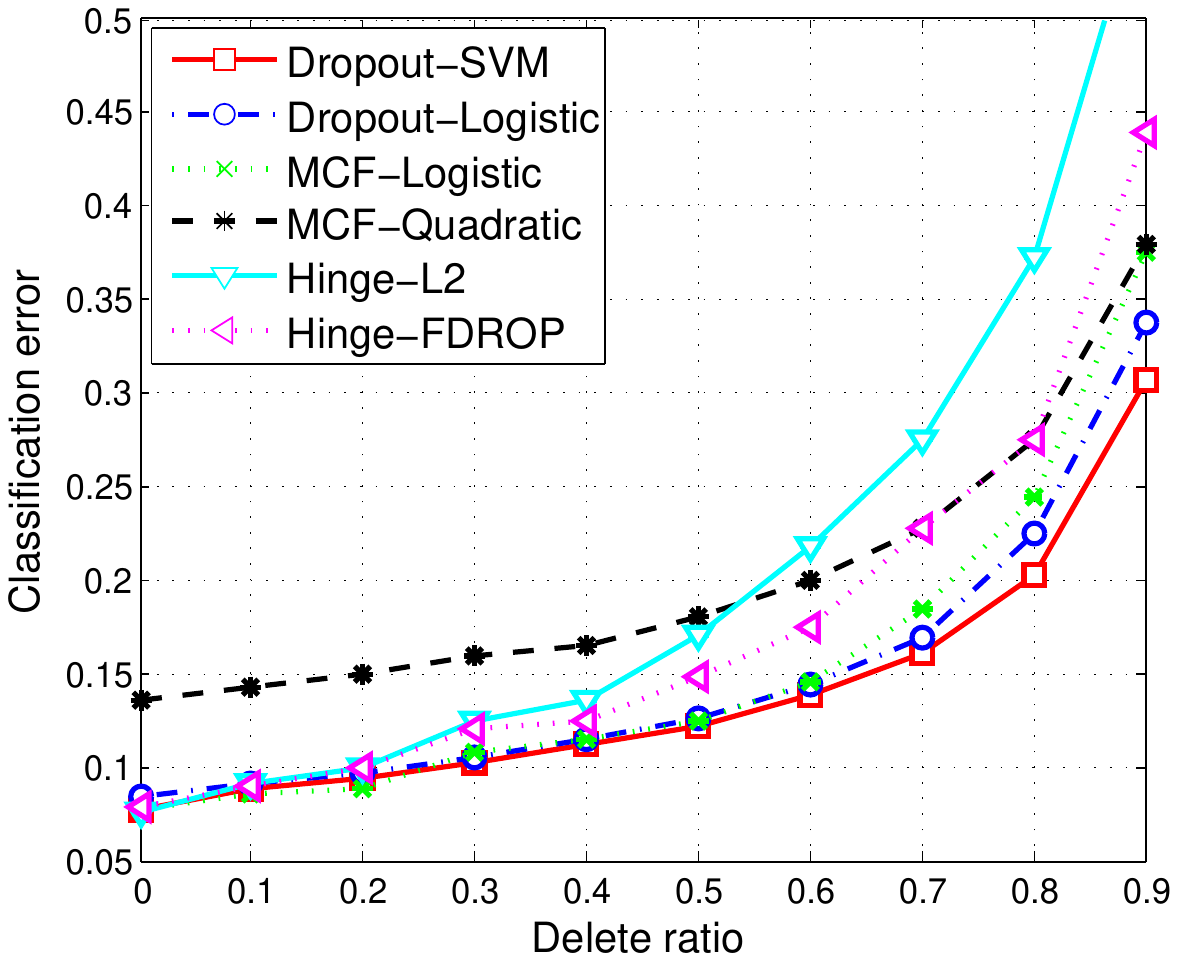}\vspace{-.2cm}
\caption{Classification errors of nightmare at test time on MNIST dataset.}\vspace{-.3cm}
\label{fig:mnist}
\end{figure}

Figure~\ref{fig:mnist} shows the classification errors of different methods as a function of the random deletion percentage of features at the testing time. Following previous settings, for each deletion percentage, we use a small validation set with the same deletion level to determine the regularization parameters and the dropout level $q$ on the whole training data. From the results, we can see that the proposed Dropout-SVM is consistently more robust than all the other competitors, including the two methods to minimize the expected logistic-loss, especially when the feature deletion percentage is high (e.g., $> 50\%$). Comparing with the standard SVM (i.e., the method Hinge-L2) and the worst-case analysis of hinge loss (i.e., Hinge-FDROP), Dropout-SVM consistently boosts the performance when the deletion ratio is greater than $10\%$. As expected, Dropout-SVM also significantly outperforms the MCF method with a quadratic loss (i.e., MCF-Quadratic), which is a special case of Dropout-SVM as shown in our theory. Finally, we also note that our iterative algorithm for the logistic-loss works slightly better than the previous algorithm (i.e., MCF-Logistic) when the deletion ratio is larger than $50\%$.

\section{Conclusions}

We present dropout training for SVMs, with an iteratively re-weighted least square (IRLS) algorithm by using data augmentation techniques. Similar ideas are applied to develop a new IRLS algorithm for the dropout training of logistic regression. Our IRLS algorithms provide insights on the connection and difference among various losses in dropout learning settings. Empirical results on various tasks demonstrate the effectiveness of our approaches.

For future work, it is remained open whether the kernel trick can be incorporated in dropout learning. We are also interested in developing more efficient algorithms, e.g., online dropout learning, to deal with even larger datasets, and investigating whether Dropout-SVM can be incorporated into a deep learning architecture or learning with latent structures~\cite{Zhu:jmlr14}.

\section*{Acknowledgments}

This work is supported by National Key Project for Basic Research of China (Grant Nos: 2013CB329403, 2012CB316301), National Natural Science Foundation of China (Nos: 61305066, 61322308, 61332007), Tsinghua Self-innovation Project (Grant Nos: 20121088071) and China Postdoctoral Science Foundation Grant (Grant Nos: 2013T60117, 2012M520281).

\bibliographystyle{aaai}
\bibliography{dpmed}

\begin{thebibliography}{}

\bibitem[\protect\citeauthoryear{Blitzer, Dredze, and
  Pereira}{2007}]{AmazonReviewData}
Blitzer, J.; Dredze, M.; and Pereira, F.
\newblock 2007.
\newblock Biographies, bollywood, boom-boxes and blenders: Domain adaptation
  for sentiment classification.
\newblock In {\em Association of Computational Linguistics}.

\bibitem[\protect\citeauthoryear{Burges and Scholkopf}{1997}]{Burges1997}
Burges, C., and Scholkopf, B.
\newblock 1997.
\newblock Improving the accuracy and speed of support vector machiens.
\newblock In {\em Advances in Neural Information Processing Systems}.

\bibitem[\protect\citeauthoryear{Chen \bgroup et al\mbox.\egroup
  }{2013}]{Chen:IJCAI2013}
Chen, N.; Zhu, J.; Xia, F.; and Zhang, B.
\newblock 2013.
\newblock Generalized relational topic models with data augmentation.
\newblock In {\em International Joint Conference on Artificial Intelligence}.

\bibitem[\protect\citeauthoryear{Dekel and Shamir}{2008}]{Dekel:icml08}
Dekel, O., and Shamir, O.
\newblock 2008.
\newblock Learning to classify with missing and corrpted features.
\newblock In {\em International Conference on Machine Learning}.

\bibitem[\protect\citeauthoryear{Globerson and Roweis}{2006}]{Globerson:icml06}
Globerson, A., and Roweis, S.
\newblock 2006.
\newblock Nightmare at test time: Robust learning by feature deletion.
\newblock In {\em International Conference on Machine Learning}.

\bibitem[\protect\citeauthoryear{Globerson \bgroup et al\mbox.\egroup
  }{2007}]{Globerson:07}
Globerson, A.; Koo, T.~Y.; Carreras, X.; and Collins, M.
\newblock 2007.
\newblock Exponentiated gradient algorithms for log-linear structured
  prediction.
\newblock In {\em ICML}.

\bibitem[\protect\citeauthoryear{Hastie, Tibshirani, and
  Friedman}{2009}]{Hastie:2009}
Hastie, T.; Tibshirani, R.; and Friedman, J.
\newblock 2009.
\newblock {\em The elements of statistical learning: data mining, inference,
  and prediction}.
\newblock Springer.

\bibitem[\protect\citeauthoryear{Hinton \bgroup et al\mbox.\egroup
  }{2012}]{Hinton:dropout}
Hinton, G.; Srivastava, N.; Krizhevsky, A.; Sutskever, I.; and Salakhutdinov,
  R.
\newblock 2012.
\newblock Improving neural networks by preventing co-adaptation of feature
  detectors.
\newblock {\em arXiv:1207.0580v1, preprint}.

\bibitem[\protect\citeauthoryear{Krizhevsky}{2009}]{Cifar-10Data}
Krizhevsky, A.
\newblock 2009.
\newblock Learning multiple layers of features from tiny images.
\newblock Technical report, University of Toronto.

\bibitem[\protect\citeauthoryear{Liu and Nocedal}{1989}]{Liu:89}
Liu, D.~C., and Nocedal, J.
\newblock 1989.
\newblock On the limited memory {BFGS} method for large scale optimization.
\newblock {\em Mathematical Programming} (45):503--528.

\bibitem[\protect\citeauthoryear{Polson and Scott}{2011}]{Polson:BA11}
Polson, N.~G., and Scott, S.~L.
\newblock 2011.
\newblock {Data Augmentation for Support Vector Machines}.
\newblock {\em Bayesian Analysis} 6(1):1--24.

\bibitem[\protect\citeauthoryear{Polson, Scott, and
  Windle}{2012}]{Polson:arXiv12}
Polson, N.~G.; Scott, J.~G.; and Windle, J.
\newblock 2012.
\newblock {Bayesian Inference for Logistic Models using {Polya-Gamma} Latent
  Variables}.
\newblock {\em arXiv:1205.0310v1}.

\bibitem[\protect\citeauthoryear{Rifkin and Klautau}{2004}]{InDefense:JMLR04}
Rifkin, R., and Klautau, A.
\newblock 2004.
\newblock In defense of one-vs-all classification.
\newblock {\em Journal of Machine Learning Research} (5):101--141.

\bibitem[\protect\citeauthoryear{Rosasco \bgroup et al\mbox.\egroup
  }{2004}]{Rosasco:2004}
Rosasco, L.; Vito, E.~D.; Caponnetto, A.; Piana, M.; and Verri, A.
\newblock 2004.
\newblock Are loss functions all the same?
\newblock {\em Neural Computation} 16(5):1063--1076.

\bibitem[\protect\citeauthoryear{Teo \bgroup et al\mbox.\egroup
  }{2008}]{Teo2008}
Teo, C.; Globerson, A.; Roweis, S.; and Smola, A.
\newblock 2008.
\newblock Convex learning with invariances.
\newblock In {\em Advances in Neural Information Processing Systems}.

\bibitem[\protect\citeauthoryear{Torralba, Fergus, and
  Freeman}{2008}]{Torralba:PAMI08}
Torralba, A.; Fergus, R.; and Freeman, W.
\newblock 2008.
\newblock A large dataset for non-parametric object and scene recognition.
\newblock {\em IEEE Transaction on Pattern Analysis and Machine Intelligence}
  30(11):1958--1970.

\bibitem[\protect\citeauthoryear{van~der Maaten \bgroup et al\mbox.\egroup
  }{2013}]{CorruptICML2013}
van~der Maaten, L.; Chen, M.; Tyree, S.; and Weinberger, K.~Q.
\newblock 2013.
\newblock Learning with marginalized corrupted features.
\newblock In {\em International Conference on Machine Learning}.

\bibitem[\protect\citeauthoryear{Vapnik}{1995}]{Vapnik:95}
Vapnik, V.
\newblock 1995.
\newblock {\em The nature of statistical learning theory}.
\newblock Springer-Verlag.

\bibitem[\protect\citeauthoryear{Vincent \bgroup et al\mbox.\egroup
  }{2008}]{Vincent:2008}
Vincent, P.; Larochelle, H.; Bengio, Y.; and Manzagol, P.~A.
\newblock 2008.
\newblock Extracting and composing robust features with denoising autoencoders.
\newblock In {\em International Conference on Machine Learning}.

\bibitem[\protect\citeauthoryear{Wager, Wang, and Liang}{2013}]{PercyLiang13}
Wager, S.; Wang, S.; and Liang, P.
\newblock 2013.
\newblock Dropout training as adaptive regularization.
\newblock In {\em Advances in Neural Information Processing}.

\bibitem[\protect\citeauthoryear{Wang and Manning}{2013}]{Wang:icml13}
Wang, S., and Manning, C.
\newblock 2013.
\newblock Fast dropout training.
\newblock In {\em International Conference on Machine Learning}.

\bibitem[\protect\citeauthoryear{Wang \bgroup et al\mbox.\egroup
  }{2013}]{Wang:emnlp13}
Wang, S.; Wang, M.; Wager, S.; Liang, P.; and Manning, C.
\newblock 2013.
\newblock Feature noising for log-linear structured prediction.
\newblock In {\em Empirical Methods in Natural Language Processing}.

\bibitem[\protect\citeauthoryear{Zhu \bgroup et al\mbox.\egroup
  }{2014}]{Zhu:jmlr14}
Zhu, J.; Chen, N.; Perkins, H.; and Zhang, B.
\newblock 2014.
\newblock Gibbs max-margin topic models with data augmentation.
\newblock {\em Journal of Machine Learning Research (JMLR)} 15:1073--1110.

\end{thebibliography}

\end{document}